\pdfoutput=1
\documentclass[letterpaper]{article} 
\usepackage{aaai2027}
\usepackage[hyphens]{url}  
\usepackage{graphicx} 
\usepackage{natbib}  
\usepackage{subcaption}
\usepackage{caption} 
\usepackage{algorithm}
\usepackage{algorithmic}
\usepackage{amssymb}
\usepackage[table]{xcolor}
\usepackage{booktabs}
\usepackage{amsmath}

\title{Adaptive FastOPD: Progress-Aware Rollout Horizon Expansion for Efficient On-Policy Distillation}

\author{Qian Tan\textsuperscript{1}, 
Huaifei Liang\textsuperscript{1},
Xuanyu Zhu\textsuperscript{3}, 
Lei Jiang\textsuperscript{1}, 
Yuqiang Li\textsuperscript{2,\dag},}
\affiliations{\textsuperscript{1}University of Science and Technology of China \\
\textsuperscript{2}Shanghai Artificial Intelligence Laboratory \\
\textsuperscript{3}Shanghai Jiao Tong University \\
\texttt{tanqian@mail.ustc.edu.cn,liyuqiang@pjlab.org.cn} }

\newcommand{\best}[1]{\textbf{#1}}
\newcommand{\second}[1]{\underline{#1}}
\newcommand{\method}{Adaptive FastOPD}

\begin{document}
\nocopyright
\maketitle

\begin{abstract}
On-policy distillation (OPD) provides dense teacher supervision along student-generated trajectories, but its online rollout process incurs substantial computational cost, particularly when a few long responses delay batch completion. Existing acceleration methods typically control rollout length using fixed budgets or absolute teacher--student agreement thresholds, which may not reflect learning progress across different models and training stages. We propose \method{}, a progress-aware strategy that expands the rollout horizon only when learning near the current boundary region has plateaued and the current horizon is sufficiently utilized. The former is determined from four teacher--student signals measured relative to their values upon entering each horizon, making expansion responsive to stage-specific progress rather than a predefined step interval or an absolute threshold on the raw agreement signals, while the latter prevents a small number of long responses from triggering increases in rollout cost. Across two teacher--student pairs, \method{} achieves the highest average performance while reducing training time by 49.1--71.2\% relative to OPD 15K, and remains robust across a range of hyperparameter settings.
\end{abstract}

\section{Introduction}

On-policy distillation (OPD) has emerged as an important
post-training approach for transferring reasoning behavior from a strong
teacher model to a student. OPD samples sequences from the
current student and uses the teacher's token-level distributions on
student-generated prefixes as supervision~\cite{agarwal2024policy}. This
on-policy interaction supplies dense
token-level supervision along a reasoning trace. Despite these advantages, the online training loop remains a major efficiency bottleneck. The cost grows with the rollout horizon. Response lengths can also vary within a batch, introducing
additional inefficiency. This problem is severe when a few long responses continue
decoding after most have terminated. Since a synchronous training iteration
cannot proceed until its rollout batch is completed, these long-tail
responses can become stragglers that determine the wall-clock time of the rollout phase~\cite{shao2026beat,khan2026faster}.

To alleviate these issues, recent work has explored shortening,
progressively expanding, or selectively continuing OPD rollouts~\cite{zhang2026fast,yang2026prune,ziheng2026less,
zhang2026full,zhao2026prefix}.  FastOPD observes that useful supervision is often concentrated near the beginning of a response and progressively increases the horizon with a fixed schedule~\cite{zhang2026fast}. Prune-OPD detects local student--teacher drift and truncates supervision after an overlap-based reliability score crosses a threshold~\cite{yang2026prune}. Early Stopping Rollout (ESR) argues that teacher supervision becomes
less corrective at later positions and therefore restricts rollout generation to a short predefined rollout horizon~\cite{ziheng2026less}. Collectively, these
studies show that truncating or selectively extending rollouts
can accelerate OPD while matching the performance of full-rollout training.

\begin{figure*}[t]
    \centering
    \includegraphics[width=0.9\textwidth]{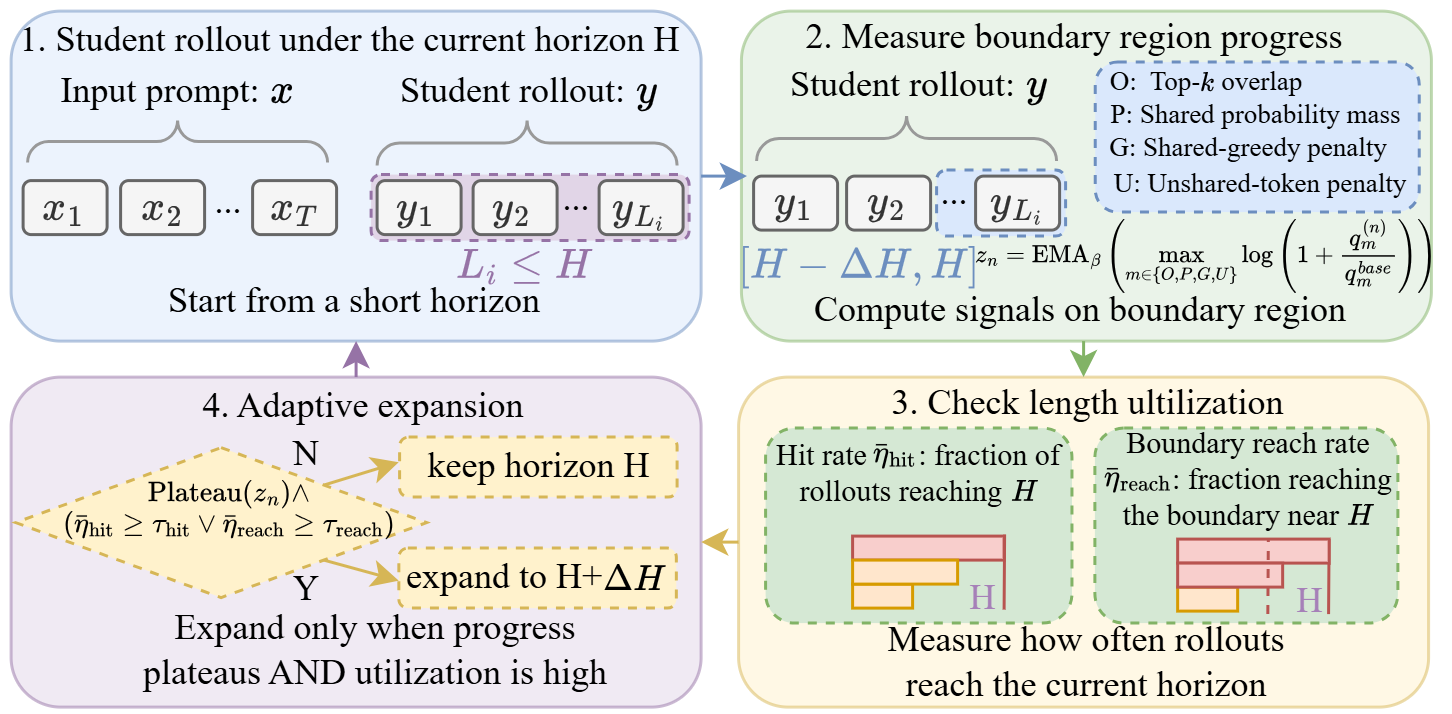}
    \caption{
Overview of Adaptive FastOPD. The student first generates responses
under the current horizon $H$, defined as the maximum response
length at the current stage. Within the current boundary region, we
measure four complementary OPD signals: top-$k$ overlap ($O$), shared
probability mass ($P$), shared-greedy penalty ($G$), and unshared-token
penalty ($U$). These signals are normalized against the baselines established when the current horizon is first entered, aggregated through the maximum normalized badness score, and smoothed with an exponential moving average to detect the plateau. In parallel, the horizon-hit rate and
boundary-reach rate verify whether the current horizon is sufficiently
utilized. The horizon is expanded from $H$ to $H+\Delta H$ only when
progress has plateaued and at least one utilization criterion is
satisfied.
}
\end{figure*}

Despite these advances, existing rollout-length adaptation methods remain sensitive to
hyperparameters that directly determine rollout expansion or
truncation. Fixed-truncation methods require the cutoff length to be
selected in advance~\cite{ziheng2026less}, while FastOPD~\cite{zhang2026fast} expands the rollout horizon at a fixed interval. Prune-OPD relies on an overlap threshold $\gamma$ to
control reward pruning and response-length adaptation~\cite{yang2026prune}. Prefix-Guided OPD (PG-OPD) instead requires a continuation budget
that determines how many candidate trajectories receive long rollouts~\cite{zhao2026prefix}. The appropriate values of these control
parameters may differ across models, datasets, and
training stages. Applying the methods to a new setting may require new hyperparameter searches, whose additional computational
cost can partially offset the intended acceleration, or suboptimal settings may degrade downstream performance. These limitations motivate the development of a more adaptive and flexible approach.

To this end, we propose \method{}, a progress-aware
rollout horizon expansion strategy for efficient on-policy distillation.
Starting from a short rollout horizon, \method{} continuously monitors learning progress in the boundary region and expands the horizon only when this progress has plateaued and the current horizon is utilized. Instead of comparing raw teacher--student signals against
a globally fixed target, \method{} establishes a stage-specific
baseline upon entering each horizon and measures subsequent changes
relative to the initial state. This self-referenced design reduces reliance on manually specified
absolute thresholds for the raw teacher--student signals.
Specifically, within the current boundary region, we measure
four complementary signals and convert these
signals into normalized badness scores, conservatively aggregate them using their maximum, and smooth the resulting aggregate badness score with an exponential moving average. The horizon is expanded only after the smoothed score has remained on a plateau for several consecutive updates, and the current horizon is sufficiently utilized, as measured by the horizon-hit rate
or boundary-reach rate. This utilization gate ensures that the progress estimate for the
boundary region is supported by enough rollouts and prevents horizon expansion when only a
few long responses would dominate batch completion time, thereby reducing end-to-end rollout time~\cite{shao2026beat,khan2026faster}. Motivated by prior findings that teacher supervision is more reliable on
shorter prefixes and becomes less corrective as prefix
drift accumulates
~\cite{li2026rethinking,ziheng2026less,zhang2026fast}, we monitor
training progress only in the current boundary region and expand horizon gradually, reducing the risk that errors accumulated in insufficiently
aligned prefixes weaken teacher supervision at later positions.
Experiments on DeepSeek-R1-Distill-Qwen-1.5B~\cite{guo2025deepseek} and Qwen3-1.7B-Base~\cite{yang2025qwen3} show that \method{} achieves better average performance than
vanilla OPD and fixed-schedule FastOPD with shorter end-to-end training
time, while showing stable performance across a range of adaptation
hyperparameters.

Our contributions can be summarized as:
\begin{itemize}
    \item We propose \method{}, a progress-aware rollout-horizon
    adaptation framework, allowing OPD to allocate rollout
    computation according to the student's evolving learning dynamics.

    \item We develop a robust horizon-expansion criterion that jointly
    considers multi-signal learning progress and boundary utilization,
    enabling reliable length transitions while avoiding inefficient
    expansion caused by sparsely occurring long rollouts.

    \item Experiments across two teacher--student pairs demonstrate that
    \method{} improves the accuracy--efficiency trade-off
    over vanilla OPD and fixed-schedule FastOPD, while maintaining stable performance across the tested adaptation settings.
\end{itemize}

\section{Related Work}
\paragraph{On-policy distillation and long-tail rollout inefficiency.}

On-policy distillation (OPD) samples trajectories from the current
student policy and queries the teacher for token-level predictive
distributions along these trajectories. GKD provides a general
formulation for autoregressive models with multiple divergence
objectives, while MiniLLM studies reverse-KL distillation under the
student-induced distribution~\cite{agarwal2024policy,gu2024minillm}. Subsequent analyses show that OPD
learning is largely driven by high-probability tokens shared by the
teacher and student~\cite{li2026rethinking,luo2026demystifying}. A major cost of OPD comes from online rollout generation. Responses in
the same batch may have different lengths, creating
uneven generation times. This inefficiency is especially severe when
most responses finish early while a small number continue for much
longer.
Inference and RL systems alleviate rollout inefficiency through techniques such as continuous batching~\cite{yu2022orca,kwon2023efficient,sheng2025hybridflow,
shao2026beat,khan2026faster}. However, such system-level techniques
cannot fully remove the straggler effect, since rollout completion may
still be dominated by the longest unfinished sequences.

\paragraph{Efficient and adaptive OPD.}

Recent methods reduce OPD cost by shortening, pruning, or adaptively
allocating rollout computation. FastOPD progressively increases the
rollout horizon according to a predefined prefix-to-suffix schedule,
motivated by the observation that short prefixes retain much of the
benefit of full-horizon OPD~\cite{zhang2026fast}. ESR uses a fixed prefix
budget, while other studies explore partial-rollout supervision and
short-to-long training
~\cite{ziheng2026less,zhang2026full,zhang2026shortopd}.
Prune-OPD truncates individual trajectories based on local top-$k$
agreement, whereas PG-OPD allocates longer continuations
using prefix-level teacher--student agreement
~\cite{yang2026prune,zhao2026prefix}. Lightning OPD instead avoids
online teacher serving by precomputing teacher targets
~\cite{wu2026lightning}. ADWIN uses delayed full-rollout probes to
adjust subsequent windows according to update admissibility
~\cite{liang2026adwin}. In contrast to methods that trigger expansion or truncation directly from a fixed rollout budget, a predefined step interval, or an absolute agreement threshold, \method{} determines horizon expansion from stage-relative training progress observed at the current horizon.

\section{Method}

\subsection{Preliminaries: On-Policy Distillation}

Let $x_i\sim\mathcal{D}$ denote a prompt, $\pi_{\theta}$ the trainable
student policy, $\pi_{\bar{\theta}}$ the frozen student policy used to
generate the current rollout batch, and $\pi_T$ the teacher policy.
Given a maximum rollout horizon $H$, OPD samples a student response
$y_i=(y_{i,0},\ldots,y_{i,L_i})
\sim\pi_{\bar{\theta}}(\cdot\mid x_i)$ with $L_i\le H$.
Each valid response position $j$ defines a student-visited state
$s_{i,j}=(x_i,y_{i,<j})$. Following the top-$k$ OPD configuration used in our experiments, we
construct the candidate set
$S_{i,j}=\operatorname{TopK}_k
(\pi_{\bar{\theta}}(\cdot\mid s_{i,j}))$
from the student distribution. For each 
$v\in S_{i,j}$, we compute student log-probability
$\ell^{S}_{i,j,v}=\log\pi_{\bar{\theta}}(v\mid s_{i,j})$ and the corresponding
teacher log-probability
$\ell^{T}_{i,j,v}=\log\pi_T(v\mid s_{i,j})$. The student is then optimized with the token-level clipped OPD objective
\begin{equation}
\mathcal{L}_{\mathrm{OPD}}(\theta)
=
\frac{1}{\sum_{i,j}\mu_{i,j}}
\sum_{i,j}\mu_{i,j}
\sum_{v\in S_{i,j}}
\ell_{\mathrm{clip}}
\left(
\rho_{i,j,v}(\theta),
A_{i,j,v}
\right),
\end{equation}
where $\mu_{i,j}$ is the response mask,
$
\rho_{i,j,v}(\theta)
=
\exp\left(
\log\pi_{\theta}(v\mid s_{i,j})
-
\ell^{S}_{i,j,v}
\right),
$
is the policy ratio, and
$
A_{i,j,v}
=
w_{i,j,v}
\left(
\ell^{T}_{i,j,v}
-
\ell^{S}_{i,j,v}
\right),
w_{i,j,v}
=
\frac{\exp(\ell^{S}_{i,j,v})}
{\sum_{u\in S_{i,j}}\exp(\ell^{S}_{i,j,u})}
$
is the candidate-level advantage with student-probability
weighting. Here, $\ell_{\mathrm{clip}}$ denotes the clipped policy surrogate used by the
underlying OPD implementation.

Vanilla OPD uses the full
rollout horizon $H_{\max}$ throughout training. FastOPD instead starts
from a short horizon $H_0$ and increases it by $\Delta H$ after a fixed interval of $F$ steps:
\begin{equation}
H_n
=
\min\left\{
H_{\max},
H_0+
\left\lfloor
\frac{n-1}{F}
\right\rfloor
\Delta H
\right\},
\label{eq:fixed_schedule}
\end{equation}
where $n$ denotes the optimization step, $F$ is the predefined expansion
interval, and $\Delta H$ is the length increment. Although this schedule reduces the cost of early rollouts, its fixed
expansion interval implicitly assumes that all horizons require the same
number of optimization steps, which may be inconsistent with the
student's learning dynamics.

\subsection{\method{}}

\method{} replaces the fixed expansion in
Equation~(\ref{eq:fixed_schedule}) with a progress-aware horizon-expansion
rule. Starting from a short horizon $H_0$, the method expands the
current horizon by one fixed-size chunk only when two conditions are
simultaneously satisfied:

\begin{enumerate}
    \item the teacher--student signals at the current boundary region have
    stopped improving; and
    \item the current boundary or horizon is reached by a sufficient fraction of
    rollouts.
\end{enumerate}

Let $\mathsf{Plateau}_n$ denote the progress condition and
$\mathsf{Use}_n$ the length-utilization condition. The horizon update is
\begin{equation}
H_{n+1}
=
\begin{cases}
\min\{H_n+\Delta H,H_{\max}\},
&
\mathsf{Plateau}_n
\land
\mathsf{Use}_n
\\
H_n,
&
\text{otherwise},
\end{cases}
\label{eq:adaptive_update}
\end{equation}

Thus, \method{} determines the duration of each horizon from observed
training progress and evaluates its signals relative to their
stage-specific initial values, avoiding a fixed step interval or an absolute threshold on
the raw teacher--student signals while reducing sensitivity to
model- and dataset-dependent signal scales. We next detail the two conditions that jointly govern horizon expansion.

\subsubsection{Stage-Relative Progress and Plateau Detection}

At optimization step $n$, let $H=H_n$ denote the current rollout horizon. We monitor its final chunk,
which we refer to as the boundary region:
\begin{equation}
\mathcal{B}_H=[H-\Delta H,H),
\end{equation}
where the monitored chunk size is identical to the horizon increment
$\Delta H$. Motivated by prior work~\cite{li2026rethinking,ziheng2026less,zhang2026fast} which shows that teacher supervision
becomes less reliable at later positions as small mismatches accumulate
and shift student prefixes away from trajectories that the teacher would
naturally follow, we monitor only the most fragile boundary region and expand
the horizon progressively so as to reduce unreliable teacher
signals on unaligned student-generated prefixes.

Specifically, at each optimization step $n$, our implementation evaluates four signals. We define these signals below. For clarity, we omit the optimization-step superscript $(n)$. Let $\mathcal{V}_H=\{(i,j)\mid \mu_{i,j}=1,\,
j\in\mathcal{B}_H\}$ denote the valid response positions within the
boundary region, where $\mu_{i,j}$ is the response mask at position $j$ of
rollout $i$. Progress statistics are updated only at optimization steps satisfying
$|\mathcal{V}_H|>0$; otherwise, the previous monitoring state is
retained. For each $(i,j)\in\mathcal{V}_H$, let
$S_{i,j}$ and $T_{i,j}=\operatorname{TopK}_k
(\pi_{T}(\cdot\mid s_{i,j}))$ denote the student and teacher top-$k$ candidate
sets at state $s_{i,j}$, respectively.

\paragraph{Top-$k$ overlap ($O$).}
We measure the fraction of student top-$k$ candidates that also appear
in the teacher top-$k$ set:
\begin{equation}
O_H
=
\frac{1}{|\mathcal{V}_H|k}
\sum_{(i,j)\in\mathcal{V}_H}
\sum_{v\in S_{i,j}}
\mathbf{1}[v\in T_{i,j}].
\end{equation}
Here, $\mathbf{1}[\cdot]$ denotes the indicator function, which equals
$1$ when the condition holds and $0$ otherwise. A larger $O_H$ indicates greater agreement between the student and
teacher candidate sets.

\paragraph{Shared probability mass ($P$).}
Let $I_{i,j}=S_{i,j}\cap T_{i,j}$ and define
$
\mathcal{V}_H^{I}
=
\left\{
(i,j)\in\mathcal{V}_H:
I_{i,j}\neq\varnothing
\right\}.
$ When $\mathcal{V}_H^{I}$ is empty, we skip the current progress
observation and retain the existing monitoring state.
We first compute the average student
and teacher probability mass assigned to the shared candidates:
\begin{align}
P_H^{S}
&=
\frac{1}{|\mathcal{V}_H^{I}|}
\sum_{(i,j)\in\mathcal{V}_H^{I}}
\sum_{v\in I_{i,j}}
\pi_{\bar{\theta}}(v\mid s_{i,j}),
\\
P_H^{T}
&=
\frac{1}{|\mathcal{V}_H^{I}|}
\sum_{(i,j)\in\mathcal{V}_H^{I}}
\sum_{v\in I_{i,j}}
\pi_T(v\mid s_{i,j}).
\end{align}
The shared probability-mass signal is
\begin{equation}
P_H=\min\{P_H^{S},P_H^{T}\}.
\end{equation}
This signal complements top-$k$ overlap by measuring whether the shared
candidates also receive high probability under both policies.

\paragraph{Shared-greedy penalty ($G$).}
For each position with $I_{i,j}\neq\varnothing$, let
$
v_{i,j}^{*}
=
\arg\max_{v\in I_{i,j}}
\pi_{\bar{\theta}}(v\mid s_{i,j})
$
denote the student's highest-probability shared candidate. Using the candidate-level OPD advantage $A_{i,j,v}$, we compute
\begin{equation}
G_H
=
\max\left\{
0,
-
\frac{1}{|\mathcal{V}_H^{I}|}
\sum_{(i,j)\in\mathcal{V}_H^{I}}
A_{i,j,v_{i,j}^{*}}
\right\}.
\end{equation}
$G_H$ examines the OPD training signal on
the student's highest-probability shared candidate. A smaller $G_H$
indicates better alignment.

\paragraph{Unshared-token penalty ($U$).}
Let
$
\Omega_H
=
\left\{
(i,j,v):
(i,j)\in\mathcal{V}_H,\;
v\in S_{i,j}\setminus T_{i,j}
\right\}
$
denote the student-only candidates. Their remaining
negative advantage is summarized as
\begin{equation}
U_H
=
\max\left\{
0,
-
\frac{1}{|\Omega_H|}
\sum_{(i,j,v)\in\Omega_H}
A_{i,j,v}
\right\}.
\end{equation}
When $\Omega_H$ is empty, we set $U_H=0$. $U_H$ captures the remaining negative OPD signal on candidates selected
only by the student. A smaller $U_H$ indicates better alignment.

\begin{table*}[t]
    \centering
    \small
    \begin{tabular}{l|cccccccc}
        \toprule
        \textbf{Method} & \textbf{AIME25} & \textbf{AIME24} & \textbf{AMC23} & \textbf{MATH-500} & \textbf{Minerva} & \textbf{Olympiad} & \textbf{Avg} & \textbf{Time} \\
        \midrule
        \multicolumn{9}{l}{\textit{DeepSeek-R1-Distill-Qwen-1.5B / JustRL-DeepSeek-1.5B}} \\
        \quad Base (Instruct) & 23.3 & 26.2 & 71.0 & 83.0 & 28.0 & 44.7 & 46.0 & -- \\
        \quad + OPD 7K & \best{34.7} & \best{46.2} & 83.7 & 86.0 & \second{33.2} & \second{52.1} & \second{56.0} & 7h32min \\
        \quad + OPD 15K & 33.9 & 43.5 & 84.6 & \best{86.2} & \best{33.5} & \second{52.1} & 55.6 & 12h18min \\
        \quad + FastOPD (Fixed) & 33.8 & 42.7 & \best{86.3} & 85.8 & \second{33.2} & \best{52.5} & 55.7 & \second{7h14min} \\
        \quad + Adaptive FastOPD & \second{34.5} & \second{45.2} & \second{85.9} & \second{86.1} & 32.8 & 51.8 & \best{56.1} & \best{6h16min} \\
        \midrule
        \multicolumn{9}{l}{\textit{Qwen3-1.7B-Base / Qwen3-8B-Base}} \\
        \quad Base & 1.4 & 1.6 & 13.7 & 25.5 & 7.5 & 9.8 & 9.9 & -- \\
        \quad + OPD 7K & \second{3.9} & \best{5.8} & \second{27.9} & \second{47.7} & \best{12.4} & \best{21.8} & \second{19.9} & \second{4h43min} \\
        \quad + OPD 15K & 3.3 & 4.5 & 25.4 & \best{47.9} & 11.9 & \second{21.6} & 19.1 & 9h05min \\
        \quad + FastOPD (Fixed) & 3.5 & 5.0 & 27.1 & 47.5 & 11.9 & 21.1 & 19.4 & 4h58min \\
        \quad + Adaptive FastOPD & \best{4.0} & \second{5.4} & \best{30.0} & 47.6 & \second{12.3} & 21.3 & \best{20.1} & \best{2h37min} \\
        \bottomrule
    \end{tabular}
    \caption{Main results on mathematical reasoning benchmarks (Avg@16). Adaptive FastOPD achieves the highest average score with the shortest training time for both model pairs. Bold and underlined values denote the best and second-best results within each model pair. Time is total wall-clock training time.}
    \label{tab:main_result}
\end{table*}

\begin{table*}[t]
    \centering
    \small
    \setlength{\tabcolsep}{2.85pt}
    \begin{tabular}{lcc|cccccccc}
        \toprule
        \textbf{Method} & \textbf{Length Inc.} & \textbf{Step Interval} & \textbf{AIME25} & \textbf{AIME24} & \textbf{AMC23} & \textbf{MATH-500} & \textbf{Minerva} & \textbf{Olympiad} & \textbf{Avg} & \textbf{Time} \\
        \midrule
        \multicolumn{11}{l}{\textit{DeepSeek-R1-Distill-Qwen-1.5B / JustRL-DeepSeek-1.5B}} \\
        \quad + FastOPD (Fixed) & 1024 & 5 & \second{33.9} & 41.4 & 83.4 & 85.7 & \second{33.0} & \second{52.1} & 54.9 & 10h45min \\
        \quad + FastOPD (Fixed) & 1024 & 10 & 23.9 & 28.3 & 71.2 & 83.3 & 28.0 & 44.9 & 46.6 & 8h14min \\
        \quad + FastOPD (Fixed) & 1024 & 20 & 33.8 & \second{42.7} & \best{86.3} & \second{85.8} & \best{33.2} & \best{52.5} & \second{55.7} & \second{7h14min} \\
        \quad + Adaptive FastOPD & 1024 & Adaptive & \best{34.5} & \best{45.2} & \second{85.9} & \best{86.1} & 32.8 & 51.8 & \best{56.1} & \best{6h16min} \\
        \midrule
        \multicolumn{11}{l}{\textit{Qwen3-1.7B-Base / Qwen3-8B-Base}} \\
        \quad + FastOPD (Fixed) & 1024 & 5 & \best{4.1} & 4.5 & \second{27.3} & \best{48.4} & 11.5 & \best{21.4} & \second{19.5} & 8h07min \\
        \quad + FastOPD (Fixed) & 1024 & 10 & 1.4 & 1.4 & 13.2 & 24.7 & 7.5 & 10.0 & 9.7 & \best{2h19min} \\
        \quad + FastOPD (Fixed) & 1024 & 20 & 3.5 & \second{5.0} & 27.1 & 47.5 & \second{11.9} & 21.1 & 19.4 & 4h58min \\
        \quad + Adaptive FastOPD & 1024 & Adaptive & \second{4.0} & \best{5.4} & \best{30.0} & \second{47.6} & \best{12.3} & \second{21.3} & \best{20.1} & \second{2h37min} \\
        \bottomrule
    \end{tabular}
    \caption{Ablation comparing \method{} with fixed-step rollout expansion. Length Inc. is the number of tokens added per expansion, and Step Interval is the number of training steps between expansions. Fixed-schedule FastOPD is sensitive to the chosen interval, with performance variation across settings.}

    \label{tab:fixed_step_ablation}
\end{table*}

We then convert the four signals into badness scores:
\begin{equation}
q_O=1-O_H,\quad
q_P=1-P_H,\quad
q_G=G_H,\quad
q_U=U_H.
\label{eq:tail_terms}
\end{equation}
For each quantity, a lower value indicates better
alignment. The scale of each signal may vary across model pairs, datasets, and
horizon stages. We therefore evaluate each metric relative to its own
state at the beginning of the current horizon. Let $\mathcal{M}=\{O,P,G,U\}$ denote the set of monitored metrics.
For every metric $m\in\mathcal{M}$, we collect the first
$N_{\mathrm{base}}$ valid boundary observations and define its
stage-specific baseline:
\begin{equation}
q_m^{\mathrm{base}}
=
\max\left\{
\epsilon_s,
\frac{1}{N_{\mathrm{base}}}
\sum_{r=1}^{N_{\mathrm{base}}}q_m^{(r)}
\right\},
\end{equation}
where $\epsilon_s>0$ avoids division by
zero. At training
step $n$, the raw badness $q_m^{(n)}$ is normalized as
\begin{equation}
b_m^{(n)}
=
\log\left(
1+\frac{q_m^{(n)}}{q_m^{\mathrm{base}}}
\right).
\end{equation}
Its evolution across training steps is used to track training progress at the current horizon: a continued decrease indicates ongoing improvement, whereas the absence of a new minimum over successive observations indicates that progress has plateaued. The logarithm preserves the ordering while
reducing the numerical effect of unusually large values, making the
subsequent smoothing and plateau detection less sensitive to transient spikes.

We then aggregate the active metrics using
\begin{equation}
b^{(n)}
=
\max_{m\in\mathcal{M}}b_m^{(n)}.
\label{eq:badscore}
\end{equation}
The maximum defines the aggregate score by the largest normalized
badness at each optimization step, so progress tracking is governed by
the least aligned metric.

The aggregate score is smoothed using an exponential moving average:
\begin{equation}
z_n
=
\beta z_{n-1}
+
(1-\beta)b^{(n)}.
\end{equation}
Let $z_n^{*}$ denote the best smoothed score observed at the current
horizon. After each horizon expansion, all monitoring statistics are reset for
the newly exposed boundary chunk. Upon receiving the first valid
progress observation at step $n$, we initialize $z_n=b^{(n)}$ and set
$z_n^{*}=z_n$. If $z_n<z_{n-1}^{*}$, we update $z_n^{*}$ and reset the
plateau counter $c_n=0$; otherwise, the counter is incremented. The
progress condition is
\begin{equation}
\mathsf{Plateau}_n
=
\mathbf{1}
\left[
c_n\ge K_{\mathrm{pat}}
\right],
\label{eq:plateau_condition}
\end{equation}
where $K_{\mathrm{pat}}$ is the the required number of consecutive
non-improving observations.

This stage-relative and progress-aware criterion allows different horizons to receive
different numbers of optimization steps according to their observed
learning progress.

\subsubsection{Length-Utilization Gate}

\begin{table*}[t]
    \centering
    \small
    \begin{tabular}{ccccc|cccccccc}
        \toprule
        \textbf{$N_{\mathrm{base}}$} & \textbf{$\beta$} & \textbf{$K_{\mathrm{pat}}$} & \textbf{$\tau_{\mathrm{hit}}$} & \textbf{$\tau_{\mathrm{reach}}$} & \textbf{AIME25} & \textbf{AIME24} & \textbf{AMC23} & \textbf{MATH-500} & \textbf{Minerva} & \textbf{Olympiad} & \textbf{Avg} & \textbf{Time} \\
        \midrule
        \multicolumn{13}{l}{\textit{DeepSeek-R1-Distill-Qwen-1.5B / JustRL-DeepSeek-1.5B}} \\
        2 & 0.8 & 7 & 0.1 & 0.3 & \best{34.7} & 43.3 & 83.4 & \second{86.2} & 32.7 & 52.0 & 55.4 & 5h13min \\
        3 & 0.6 & 7 & 0.1 & 0.3 & \best{34.7} & \second{46.0} & 83.1 & \best{86.6} & 33.1 & \best{52.2} & \second{55.9} & 5h54min \\
        3 & 0.8 & 5 & 0.1 & 0.3 & \second{34.5} & 45.2 & \second{85.9} & 86.1 & 32.8 & 51.8 & \best{56.1} & 6h16min \\
        3 & 0.8 & 7 & 0.1 & 0.3 & 33.9 & \best{46.8} & 84.0 & 86.0 & 32.7 & \second{52.1} & \second{55.9} & 5h31min \\
        3 & 0.8 & 10 & 0.1 & 0.3 & 33.3 & 45.2 & \best{86.0} & 86.0 & \second{33.2} & 51.6 & \second{55.9} & \best{4h55min} \\
        3 & 0.8 & 7 & 0.0 & 0.0 & 33.1 & 43.7 & 83.7 & 85.8 & \best{33.3} & 51.8 & 55.2 & \second{5h00min} \\
        \midrule
        \multicolumn{13}{l}{\textit{Qwen3-1.7B-Base / Qwen3-8B-Base}} \\
        2 & 0.8 & 7 & 0.1 & 0.3 & \best{4.6} & \best{6.3} & 28.4 & \second{48.6} & 12.2 & 21.7 & \best{20.3} & \second{2h40min} \\
        3 & 0.6 & 7 & 0.1 & 0.3 & \second{4.0} & \second{6.0} & 28.3 & \best{48.8} & 11.7 & 21.7 & \second{20.1} & 2h47min \\
        3 & 0.8 & 5 & 0.1 & 0.3 & \second{4.0} & 5.4 & \best{30.0} & 47.6 & \second{12.3} & 21.3 & \second{20.1} & \best{2h37min} \\
        3 & 0.8 & 7 & 0.1 & 0.3 & 3.8 & 5.8 & 28.1 & 48.3 & 12.1 & 21.8 & 20.0 & 2h49min \\
        3 & 0.8 & 10 & 0.1 & 0.3 & 3.5 & 3.8 & 28.1 & 48.5 & \best{12.8} & \best{22.6} & 19.9 & 2h41min \\
        3 & 0.8 & 7 & 0.0 & 0.0 & 3.1 & 5.2 & \second{28.5} & 48.0 & \best{12.8} & \second{22.1} & 19.9 & 6h06min \\
        \bottomrule
    \end{tabular}
    \caption{Hyperparameter ablation of \method{} (Avg@16). $N_{\mathrm{base}}$ is the number of valid boundary observations at each horizon, $\beta$ is the EMA coefficient, $K_{\mathrm{pat}}$ is the plateau patience, and $\tau_{\mathrm{hit}}$ and $\tau_{\mathrm{reach}}$ are the utilization thresholds. \method{} shows limited variation in average performance across the tested settings for both model pairs.}
    \label{tab:hyperparameter_ablation}
\end{table*}

Let $L_i^{(n)}$ denote the
generated length of rollout $i$ at optimization step $n$, and let
$N_{\mathrm{roll}}$ be the number of rollouts in the batch. We define
\begin{align}
\eta_{\mathrm{hit}}^{(n)}
&=
\frac{1}{N_{\mathrm{roll}}}
\sum_{i=1}^{N_{\mathrm{roll}}}
\mathbf{1}
\left[
L_i^{(n)}\ge H
\right],
\\
\eta_{\mathrm{reach}}^{(n)}
&=
\frac{1}{N_{\mathrm{roll}}}
\sum_{i=1}^{N_{\mathrm{roll}}}
\mathbf{1}
\left[
L_i^{(n)}\ge H-\Delta H
\right].
\end{align}
The horizon-hit rate measures the fraction of rollouts that reach the current generation limit,
whereas the boundary-reach rate measures how often they enter the boundary.

We smooth both rates using
$
\bar{\eta}_{a}^{(n)}
=
\beta\bar{\eta}_{a}^{(n-1)}
+
(1-\beta)\eta_{a}^{(n)},
\quad
a\in\{\mathrm{hit},\mathrm{reach}\},
$ where we initialize
$\bar{\eta}_{a}^{(n_0)}=\eta_{a}^{(n_0)}$ for the first observation at a newly entered horizon,
and define the utilization condition:
\begin{equation}
\mathsf{Use}_n
=
\mathbf{1}
\left[
\bar{\eta}_{\mathrm{hit}}^{(n)}
\ge\tau_{\mathrm{hit}}
\;\lor\;
\bar{\eta}_{\mathrm{reach}}^{(n)}
\ge\tau_{\mathrm{reach}}
\right].
\label{eq:lengthgate}
\end{equation}
The disjunction allows the utilization condition to be satisfied either
when a sufficient fraction of rollouts approaches the generation limit
or when broader coverage of the boundary is observed.

When few rollouts reach the boundary, the progress estimate
is based on a small and potentially unrepresentative subset of the batch.
Expanding the horizon in this regime would also primarily extend a few
long-running responses, which can dominate batch completion time
~\cite{shao2026beat,khan2026faster}. Requiring sufficient boundary
coverage therefore improves the reliability of the progress estimate
and avoids length expansion that provides little additional supervision
for most rollouts.

\section{Experiments}

\subsection{Experimental Setup}

\paragraph{Training data and implementation.}
We train all methods on DAPO-Math-17K~\cite{yu2026dapo} using the verl
framework~\cite{li2026rethinking,sheng2025hybridflow}, with vLLM for
rollout generation~\cite{kwon2023efficient}. We evaluate two
teacher--student configurations: JustRL-DeepSeek-1.5B~\cite{he2025justrl}
with DeepSeek-R1-Distill-Qwen-1.5B~\cite{guo2025deepseek}, and
Qwen3-8B-Base with Qwen3-1.7B-Base~\cite{yang2025qwen3}. All experiments
run on a single node with four NVIDIA H200 140GB GPUs, with the actor
trained in FP32. Each generation batch contains 64 inputs and four
responses per input. We use a learning rate of $1\times10^{-6}$,
student-selected top-$16$ candidates with student-probability weighting,
and no additional KL regularization. All runs use one epoch without data
shuffling, corresponding to 279 optimization steps. Both adaptive and
fixed schedules start from a rollout horizon of 1,024 tokens, use
$\Delta H=1{,}024$, and allow a maximum horizon of 15,360 tokens. The
main \method{} configuration uses $N_{\mathrm{base}}=3$, $\beta=0.8$,
$K_{\mathrm{pat}}=5$, $\tau_{\mathrm{hit}}=0.1$, and
$\tau_{\mathrm{reach}}=0.3$, while fixed-schedule FastOPD expands the
horizon every 20 optimization steps.

\paragraph{Baselines and evaluation.}
We compare \method{} with the original student model (Base), OPD using
fixed rollout of 7,168 and 15,360 tokens (OPD 7K/15K), and
fixed-schedule FastOPD (FastOPD (Fixed)). The OPD baselines
characterize the performance and computational cost of different rollout
budgets, while FastOPD (Fixed) isolates the effect of replacing a
predefined expansion schedule with our progress-aware strategy. We evaluate on AIME
2025~\cite{balunovic2026matharena}, AIME 2024 and AMC
2023~\cite{yang2024qwen2}, MATH-500~\cite{hendrycks2021math}, Minerva
Math~\cite{lewkowycz2022solving}, and
OlympiadBench~\cite{he2024olympiadbench}. For each problem, we sample 16
responses with temperature $0.7$, top-$p=0.95$, and a maximum generation
length of 31,744 tokens. We report the mean exact-match correctness over
the 16 samples for each benchmark and the macro average across all six
benchmarks (Avg@16).

\begin{figure*}[t]
    \centering
    \begin{subfigure}[t]{0.33\textwidth}
        \centering
        \includegraphics[width=\linewidth]{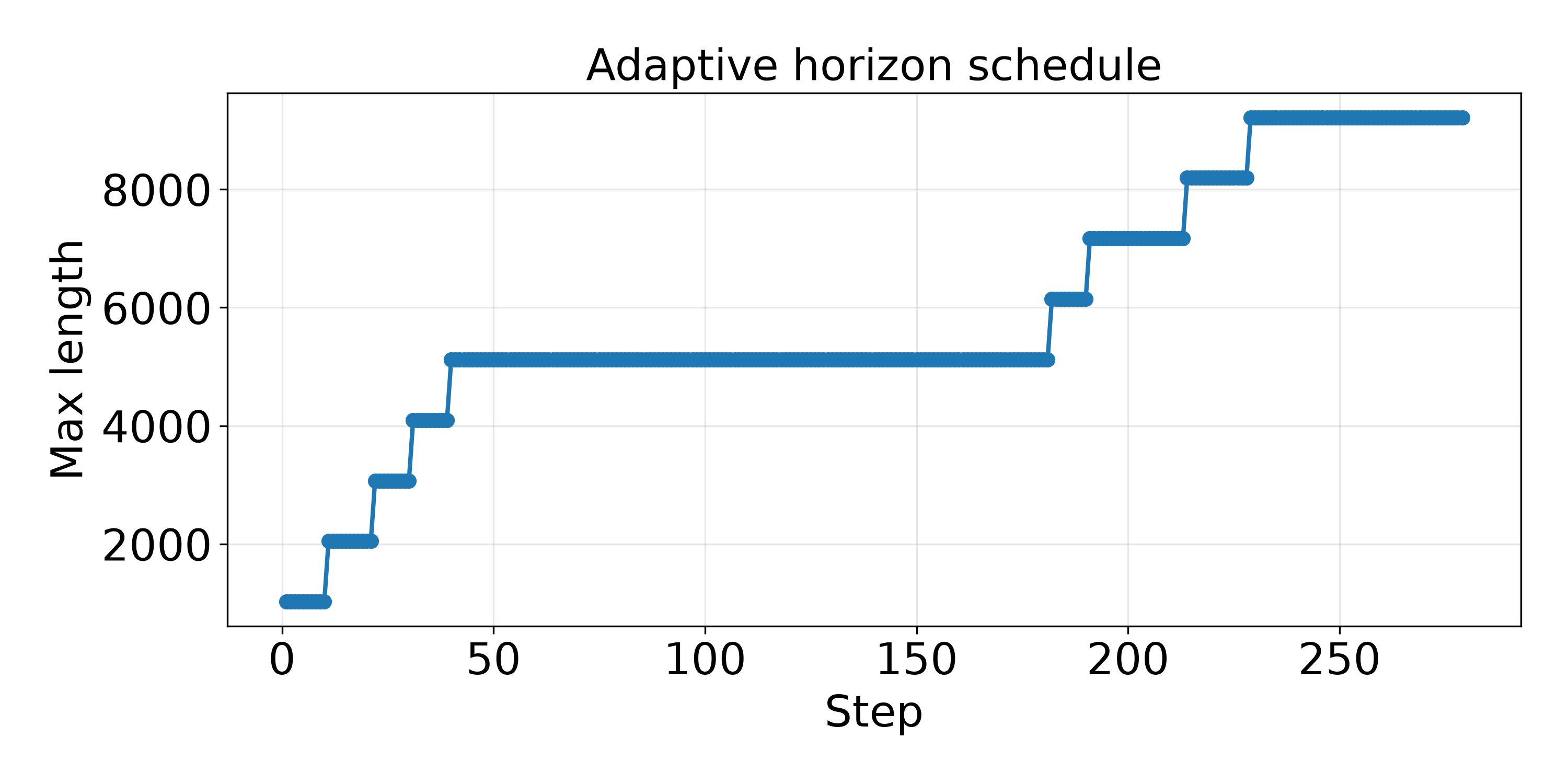}
        \caption{Adaptive horizon schedule.}
    \end{subfigure}
    \hfill
    \begin{subfigure}[t]{0.33\textwidth}
        \centering
        \includegraphics[width=\linewidth]{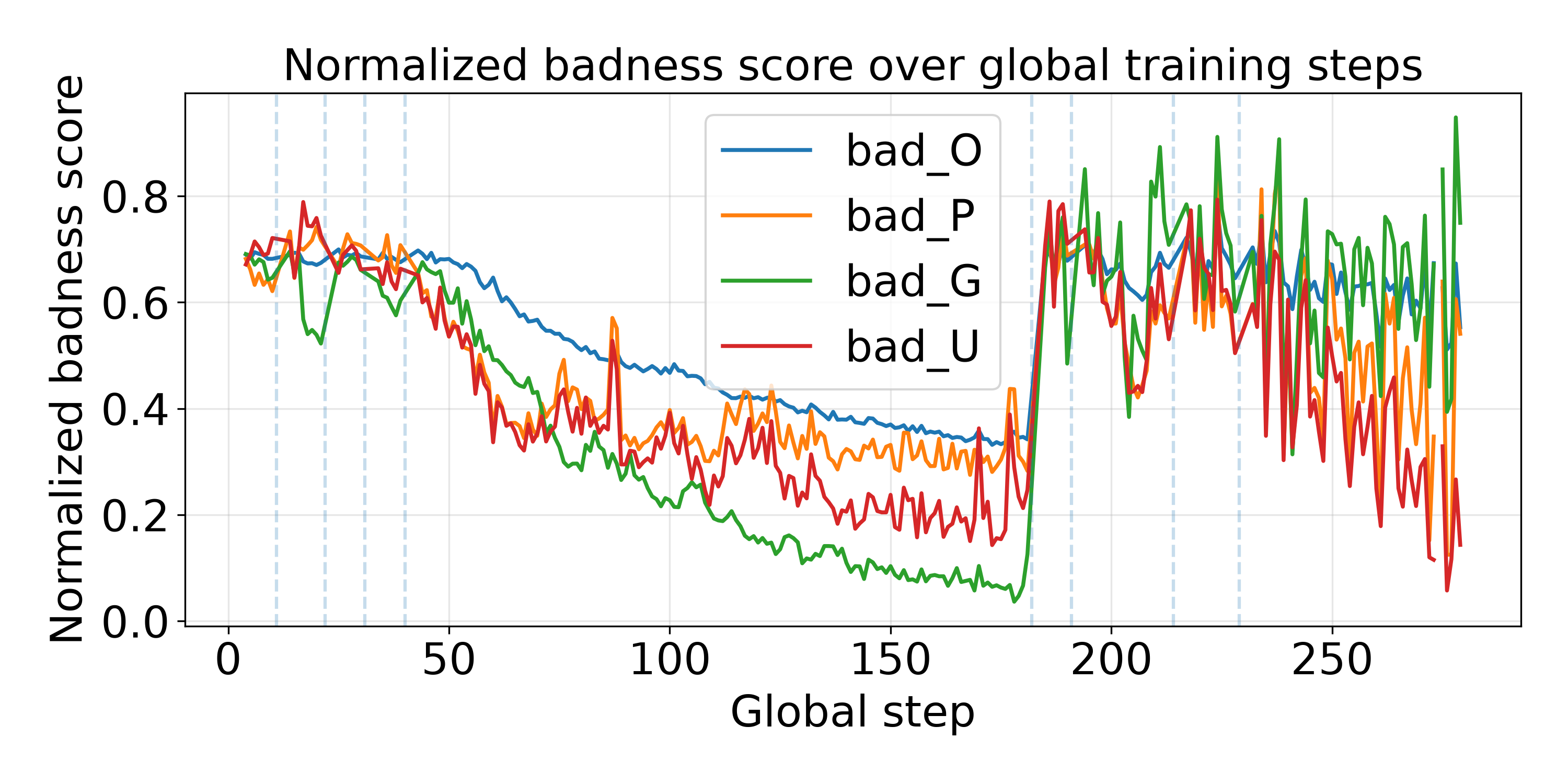}
        \caption{Normalized badness score over training.}
    \end{subfigure}
    \hfill
    \begin{subfigure}[t]{0.33\textwidth}
        \centering
        \includegraphics[width=\linewidth]{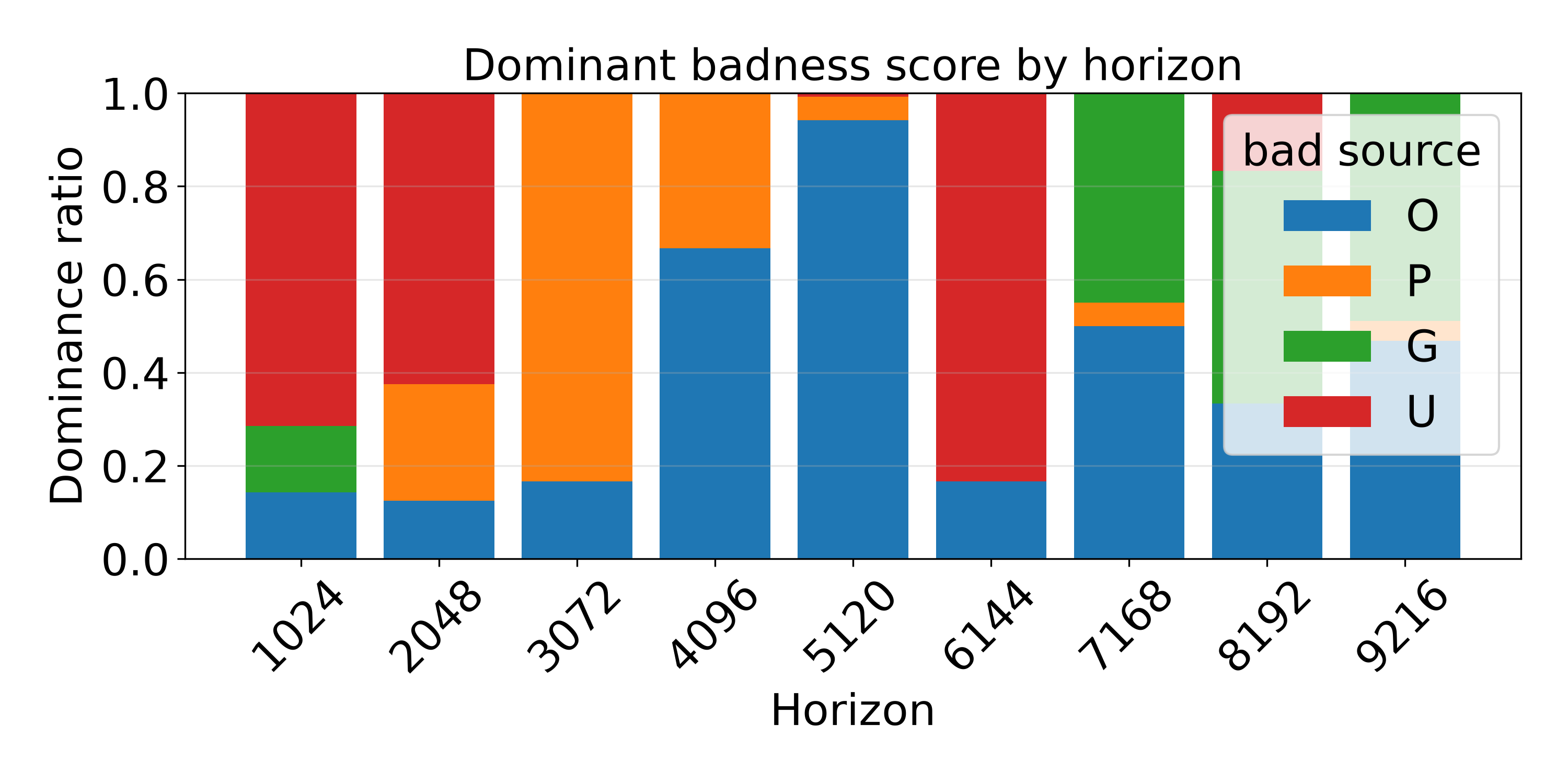}
        \caption{Dominant badness score by horizon.}
    \end{subfigure}
    \caption{Horizon-adaptation diagnostics for DeepSeek-R1-Distill-Qwen-1.5B with $(N_{\mathrm{base}},\beta,K_{\mathrm{pat}},
\tau_{\mathrm{hit}},\tau_{\mathrm{reach}})
=(3,0.8,5,0.1,0.3)$. The resulting adaptive schedule spends different numbers of updates at different horizons, and the identity of the maximum normalized badness score changes across stages.}
    \label{fig:controller_dynamics}
\end{figure*}

\begin{table*}[t]
    \centering
    \small
    \begin{tabular}{lcccccccc}
        \toprule
        \textbf{Signals} & \textbf{AIME25} & \textbf{AIME24} & \textbf{AMC23} & \textbf{MATH-500} & \textbf{Minerva} & \textbf{Olympiad} & \textbf{Avg} & \textbf{Time} \\
        \midrule
        \quad POGU & \second{34.5} & \second{45.2} & \best{85.9} & \best{86.1} & 32.8 & \best{51.8} & \best{56.1} & 6h16min \\
        \quad POG & 33.1 & 45.0 & 84.5 & \second{86.0} & \best{33.2} & \second{51.7} & 55.6 & 5h27min \\
        \quad OG & \best{35.0} & \best{46.0} & \second{84.6} & 85.4 & \second{32.9} & 51.4 & \second{55.9} & \second{3h31min} \\
        \quad O & 34.4 & 43.5 & 80.8 & 83.9 & 32.6 & 49.7 & 54.2 & \best{2h29min} \\
        \bottomrule
    \end{tabular}
    \caption{Ablation of progress signals in \method{} on DeepSeek-R1-Distill-Qwen-1.5B / JustRL-DeepSeek-1.5B (Avg@16). Using multiple signals yields a more conservative criterion of training progress than relying on overlap alone, leading to better final performance. The gray row denotes the full POGU configuration used in the main experiments.}

    \label{tab:signal_ablation}
\end{table*}

\subsection{Main Results}

Table~\ref{tab:main_result} shows that adaptive expansion improves the accuracy--time trade-off on two teacher--student configurations. On the DeepSeek pair, \method{} reaches the best average score, 56.1, in 6h16min. It exceeds OPD 7K and OPD 15K while using 16.8\% and 49.1\% less time. It also improves over the fixed-schedule FastOPD baseline by 0.4 average points with 13.4\% less time. The per-benchmark results show that \method{} remains competitive across all six benchmarks, indicating that its average advantage is not driven by a single outlier task. The efficiency gain is larger for the Qwen3-1.7B-Base student. \method{} achieves the best average, 20.1, in 2h37min, compared with 19.4 in 4h58min for fixed-schedule FastOPD and 19.1 in 9h05min for OPD 15K. This corresponds to 47.3\% and 71.2\% less training time. OPD 7K remains strong on several individual benchmarks, but the adaptive schedule obtains a higher macro average while using 44.5\% less time. These results suggest that neither a uniformly short horizon nor uniform expansion toward the maximum is consistently optimal.

\subsection{Ablation Study}
\subsubsection{Sensitivity of Fixed Schedules}

Table~\ref{tab:fixed_step_ablation} compares different expansion intervals for fixed-schedule FastOPD. For the DeepSeek pair, expanding every 20 steps is strongest among fixed schedules, while the 10-step schedule collapses from 55.7 to 46.6 average. For Qwen3, the 5- and 20-step schedules are similar in accuracy, whereas the 10-step schedule ends early and falls to 9.7. Across the tested fixed schedules, the mean score is 52.4 for DeepSeek
and 16.2 for Qwen3. The corresponding differences between the best and
worst schedules are 9.1 and 9.8 percentage points, respectively. A fixed schedule can expand before the current prefix is aligned, or spend many updates at a horizon that has already saturated. \method{} avoids choosing a global interval and it is both faster and more accurate than the strongest fixed schedule in the DeepSeek setting. In the Qwen3 setting, the fastest fixed run is 18 minutes shorter but loses 10.4 average points; the adaptive run recovers the strongest accuracy while remaining faster than the competitive fixed alternatives.

\subsubsection{Robustness and the Value of Length Utilization}

Table~\ref{tab:hyperparameter_ablation} reports model performance under different hyperparameter settings. Despite varying $N_{\mathrm{base}}$, $\beta$, $K_{\mathrm{pat}}$, $\tau_{\mathrm{hit}}$, and $\tau_{\mathrm{reach}}$, the DeepSeek and Qwen3 runs remain within narrow ranges of 55.2--56.1 and 19.9--20.3, respectively. These ranges are much smaller than those of the fixed schedules in Table~\ref{tab:fixed_step_ablation}. $N_{\mathrm{base}}$, $\beta$, and $K_{\mathrm{pat}}$ affect how quickly evidence accumulates, but the expansion decision depends on the observed training dynamics rather than a fixed absolute threshold on raw teacher--student agreement metric. The last row of each group disables both length thresholds. On DeepSeek, removing the gate lowers performance from 56.1 to 55.2
while reducing training time from 6h16min to 5h00min. On Qwen3, removing the gate both lowers performance and increases time from 2h37min to 6h06min. Inspection of the schedules shows that ungated expansions can expose expensive horizons that are reached by only a small tail of the rollout distribution. The length-utilization gate therefore improves reliability and helps prevent a small number of long rollouts from increasing training time.

\subsubsection{Why Multiple Signals}

Using the DeepSeek model as an example, Figure~\ref{fig:controller_dynamics} visualizes the actual rollout-horizon schedule during the main experiment, the evolution of the normalized badness score, and the signal that dominates the expansion decision at each horizon.
The resulting adaptive schedule is visibly non-uniform: \method{} passes quickly through some early horizons but remains for many updates at 5,120 tokens. Across the run, the total dominance duration decreases approximately in the order
OGPU and no single signal dominates every stage. Table~\ref{tab:signal_ablation} evaluates a sequence of nested signal sets constructed by progressively removing the signal with the smallest share of dominant expansion decisions. Full POGU achieves the highest average score of 56.1. Removing U, or removing both P and U, reduces training time but also lowers final performance. Using only the most frequently dominant overlap signal produces the fastest run but decreases the average score to 54.2. These results support the multi-signal aggregation in Equation~(\ref{eq:badscore}). By taking the maximum normalized badness, the expansion decision is based on the signal that currently indicates the greatest remaining teacher--student discrepancy, while the dominant signal may change across horizons and training stages. Combining complementary signals provides a more comprehensive estimate of training progress and yields better final performance than using overlap alone. The ablation also reveals a clear speed--performance trade-off. The OG variant reaches an average score of 55.9 in 3h31min, close to the 55.7 achieved by fixed-schedule FastOPD in 7h14min. The O-only variant obtains a competitive score of 54.2 using about 20\% of the training time required by OPD 15K. Thus, the adaptive strategy can favor greater efficiency by using a smaller signal set, while the full signal set is preferable when final performance is the primary objective.

\section{Conclusion}

We introduced \method{}, a progress-aware rollout-horizon expansion strategy for efficient on-policy distillation. Rather than triggering horizon expansion at a fixed step interval or
from an absolute threshold on the raw teacher--student signals, \method{} evaluates learning progress relative to the initial state of each horizon. It combines four signals to monitor optimization near the current boundary region and expands the horizon only after the aggregated progress has plateaued and the available length is sufficiently utilized. The utilization condition also avoids extending the horizon when only a small number of long responses would increase rollout time. Experiments across two teacher--student pairs show that \method{} achieves the highest average performance with lower training time than fixed-schedule FastOPD and OPD 15K. The method is also less sensitive to the tested adaptation hyperparameters, supporting progress-aware adaptation as a more robust method. Future work may investigate additional progress signals and reduce the remaining dependence on other adaptation parameters.

\bibliography{aaai2027}

\begin{thebibliography}{26}
\providecommand{\natexlab}[1]{#1}

\bibitem[{Agarwal et~al.(2024)Agarwal, Vieillard, Zhou, Stanczyk, Ramos~Garea,
  Geist, and Bachem}]{agarwal2024policy}
Agarwal, R.; Vieillard, N.; Zhou, Y.; Stanczyk, P.; Ramos~Garea, S.; Geist, M.;
  and Bachem, O. 2024.
\newblock On-policy distillation of language models: Learning from
  self-generated mistakes.
\newblock In \emph{International Conference on Learning Representations},
  volume 2024, 21246--21263.

\bibitem[{Balunovic et~al.(2026)Balunovic, Dekoninck, Petrov, Jovanovi{\'c},
  and Vechev}]{balunovic2026matharena}
Balunovic, M.; Dekoninck, J.; Petrov, I.; Jovanovi{\'c}, N.; and Vechev, M.
  2026.
\newblock Matharena: Evaluating llms on uncontaminated math competitions.
\newblock \emph{Advances in Neural Information Processing Systems}, 38.

\bibitem[{Gu et~al.(2024)Gu, Dong, Wei, and Huang}]{gu2024minillm}
Gu, Y.; Dong, L.; Wei, F.; and Huang, M. 2024.
\newblock {Minillm}: Knowledge distillation of large language models.
\newblock In \emph{The twelfth international conference on learning
  representations}.

\bibitem[{Guo et~al.(2025)Guo, Yang, Zhang, Song, Wang, Zhu, Xu, Zhang, Ma, Bi
  et~al.}]{guo2025deepseek}
Guo, D.; Yang, D.; Zhang, H.; Song, J.; Wang, P.; Zhu, Q.; Xu, R.; Zhang, R.;
  Ma, S.; Bi, X.; et~al. 2025.
\newblock DeepSeek-R1 incentivizes reasoning in LLMs through reinforcement
  learning.
\newblock \emph{Nature}, 645(8081): 633--638.

\bibitem[{He et~al.(2025)He, Qu, Liu, Chen, Zuo, Qian, Zhang, Chen, Xiao, Cui
  et~al.}]{he2025justrl}
He, B.; Qu, Z.; Liu, Z.; Chen, Y.; Zuo, Y.; Qian, C.; Zhang, K.; Chen, W.;
  Xiao, C.; Cui, G.; et~al. 2025.
\newblock {Justrl}: Scaling a 1.5 b llm with a simple rl recipe.
\newblock \emph{arXiv preprint arXiv:2512.16649}.

\bibitem[{He et~al.(2024)He, Luo, Bai, Hu, Thai, Shen, Hu, Han, Huang, Zhang
  et~al.}]{he2024olympiadbench}
He, C.; Luo, R.; Bai, Y.; Hu, S.; Thai, Z.; Shen, J.; Hu, J.; Han, X.; Huang,
  Y.; Zhang, Y.; et~al. 2024.
\newblock Olympiadbench: A challenging benchmark for promoting agi with
  olympiad-level bilingual multimodal scientific problems.
\newblock In \emph{Proceedings of the 62nd Annual Meeting of the Association
  for Computational Linguistics (Volume 1: Long Papers)}, 3828--3850.

\bibitem[{Hendrycks et~al.(2021)Hendrycks, Burns, Kadavath, Arora, Basart,
  Tang, Song, and Steinhardt}]{hendrycks2021math}
Hendrycks, D.; Burns, C.; Kadavath, S.; Arora, A.; Basart, S.; Tang, E.; Song,
  D.; and Steinhardt, J. 2021.
\newblock Measuring Mathematical Problem Solving With the MATH Dataset.
\newblock In Vanschoren, J.; and Yeung, S., eds., \emph{Proceedings of the
  Neural Information Processing Systems Track on Datasets and Benchmarks},
  volume~1.

\bibitem[{Khan et~al.(2026)Khan, Ahmed, Fayyaz, Di, Hong, and
  Anwar}]{khan2026faster}
Khan, A.~A.; Ahmed, A.; Fayyaz, Z.; Di, S.; Hong, M.; and Anwar, A. 2026.
\newblock Faster Synchronous On-Policy RL via Straggler-Aware Group Sizing.
\newblock \emph{arXiv preprint arXiv:2606.02218}.

\bibitem[{Kwon et~al.(2023)Kwon, Li, Zhuang, Sheng, Zheng, Yu, Gonzalez, Zhang,
  and Stoica}]{kwon2023efficient}
Kwon, W.; Li, Z.; Zhuang, S.; Sheng, Y.; Zheng, L.; Yu, C.~H.; Gonzalez, J.;
  Zhang, H.; and Stoica, I. 2023.
\newblock Efficient memory management for large language model serving with
  pagedattention.
\newblock In \emph{Proceedings of the 29th symposium on operating systems
  principles}, 611--626.

\bibitem[{Lewkowycz et~al.(2022)Lewkowycz, Andreassen, Dohan, Dyer,
  Michalewski, Ramasesh, Slone, Anil, Schlag, Gutman-Solo
  et~al.}]{lewkowycz2022solving}
Lewkowycz, A.; Andreassen, A.; Dohan, D.; Dyer, E.; Michalewski, H.; Ramasesh,
  V.; Slone, A.; Anil, C.; Schlag, I.; Gutman-Solo, T.; et~al. 2022.
\newblock Solving quantitative reasoning problems with language models.
\newblock \emph{Advances in neural information processing systems}, 35:
  3843--3857.

\bibitem[{Li et~al.(2026)Li, Zuo, He, Zhang, Xiao, Qian, Yu, {Huan-ang Gao},
  Yang, Liu, and Ding}]{li2026rethinking}
Li, Y.; Zuo, Y.; He, B.; Zhang, J.; Xiao, C.; Qian, C.; Yu, T.; {Huan-ang Gao};
  Yang, W.; Liu, Z.; and Ding, N. 2026.
\newblock Rethinking On-Policy Distillation of Large Language Models:
  Phenomenology, Mechanism, and Recipe.
\newblock In \emph{ICML 2026 Workshop on Foundations of Deep Generative Models:
  Understanding Memorization, Generalization, and Reasoning}.

\bibitem[{Liang et~al.(2026)Liang, Tang, Bai, Liu, Yang, and
  Wu}]{liang2026adwin}
Liang, K.; Tang, C.; Bai, C.; Liu, W.; Yang, S.; and Wu, Y. 2026.
\newblock ADWIN: Adaptive Windows for Horizon-Aware On-Policy Distillation.
\newblock \emph{arXiv preprint arXiv:2605.28396}.

\bibitem[{Luo et~al.(2026)Luo, Chuang, Wang, Xu, Han, Zhang, and
  Braverman}]{luo2026demystifying}
Luo, F.; Chuang, Y.-N.; Wang, G.; Xu, Z.; Han, X.; Zhang, T.; and Braverman, V.
  2026.
\newblock Demystifying opd: Length inflation and stabilization strategies for
  large language models.
\newblock \emph{arXiv preprint arXiv:2604.08527}.

\bibitem[{Shao et~al.(2026)Shao, Srivatsa, Srivastava, Wu, Ariyak, Wu, Patel,
  Wang, Liang, Dao, Zhang, Zhang, Athiwaratkun, Xu, and Wang}]{shao2026beat}
Shao, Z.; Srivatsa, V.; Srivastava, S.; Wu, Q.; Ariyak, A.; Wu, X.; Patel, A.;
  Wang, J.; Liang, P.; Dao, T.; Zhang, C.; Zhang, Y.; Athiwaratkun, B.; Xu, C.;
  and Wang, J. 2026.
\newblock Beat the long tail: Distribution-Aware Speculative Decoding for RL
  Training.
\newblock In \emph{Proceedings of the 9th Annual Conference on Machine Learning
  and Systems (MLSys)}.

\bibitem[{Sheng et~al.(2025)Sheng, Zhang, Ye, Wu, Zhang, Zhang, Peng, Lin, and
  Wu}]{sheng2025hybridflow}
Sheng, G.; Zhang, C.; Ye, Z.; Wu, X.; Zhang, W.; Zhang, R.; Peng, Y.; Lin, H.;
  and Wu, C. 2025.
\newblock HybridFlow: A Flexible and Efficient RLHF Framework.

\bibitem[{Wu, Han, and Cai(2026)}]{wu2026lightning}
Wu, Y.; Han, S.; and Cai, H. 2026.
\newblock Lightning opd: Efficient post-training for large reasoning models
  with offline on-policy distillation.
\newblock \emph{arXiv preprint arXiv:2604.13010}.

\bibitem[{Yang et~al.(2025)Yang, Li, Yang, Zhang, Hui, Zheng, Yu, Gao, Huang,
  Lv et~al.}]{yang2025qwen3}
Yang, A.; Li, A.; Yang, B.; Zhang, B.; Hui, B.; Zheng, B.; Yu, B.; Gao, C.;
  Huang, C.; Lv, C.; et~al. 2025.
\newblock Qwen3 technical report.
\newblock \emph{arXiv preprint arXiv:2505.09388}.

\bibitem[{Yang et~al.(2024)Yang, Zhang, Hui, Gao, Yu, Li, Liu, Tu, Zhou, Lin
  et~al.}]{yang2024qwen2}
Yang, A.; Zhang, B.; Hui, B.; Gao, B.; Yu, B.; Li, C.; Liu, D.; Tu, J.; Zhou,
  J.; Lin, J.; et~al. 2024.
\newblock Qwen2. 5-math technical report: Toward mathematical expert model via
  self-improvement.
\newblock \emph{arXiv preprint arXiv:2409.12122}.

\bibitem[{Yang et~al.(2026)Yang, Guo, Song, Xu, Wang, Wang, Liang, and
  Tang}]{yang2026prune}
Yang, Z.; Guo, Z.; Song, Y.; Xu, M.; Wang, Y.; Wang, Y.; Liang, X.; and Tang,
  J. 2026.
\newblock Prune-OPD: Efficient and Reliable On-Policy Distillation for
  Long-Horizon Reasoning.
\newblock \emph{arXiv preprint arXiv:2605.07804}.

\bibitem[{Yu et~al.(2022)Yu, Jeong, Kim, Kim, and Chun}]{yu2022orca}
Yu, G.-I.; Jeong, J.~S.; Kim, G.-W.; Kim, S.; and Chun, B.-G. 2022.
\newblock Orca: A distributed serving system for $\{$Transformer-Based$\}$
  generative models.
\newblock In \emph{16th USENIX symposium on operating systems design and
  implementation (OSDI 22)}, 521--538.

\bibitem[{Yu et~al.(2026)Yu, Zhang, Zhu, Yuan, Zuo, Yue, Dai, Fan, Liu, Liu
  et~al.}]{yu2026dapo}
Yu, Q.; Zhang, Z.; Zhu, R.; Yuan, Y.; Zuo, X.; Yue, Y.; Dai, W.; Fan, T.; Liu,
  G.; Liu, L.; et~al. 2026.
\newblock Dapo: An open-source llm reinforcement learning system at scale.
\newblock \emph{Advances in Neural Information Processing Systems}, 38:
  113222--113244.

\bibitem[{Zhang et~al.(2026{\natexlab{a}})Zhang, Yang, Janghorbani, Han,
  Ressler~II, Qian, Lyng, Batra, and Tillman}]{zhang2026fast}
Zhang, D.; Yang, Z.; Janghorbani, S.; Han, J.; Ressler~II, A.; Qian, Q.; Lyng,
  G.~D.; Batra, S.~S.; and Tillman, R.~E. 2026{\natexlab{a}}.
\newblock Fast and effective on-policy distillation from reasoning prefixes.
\newblock In \emph{Findings of the Association for Computational Linguistics:
  ACL 2026}, 25553--25569.

\bibitem[{Zhang et~al.(2026{\natexlab{b}})Zhang, Yuan, Lin, Lu, Han, Sun, Li,
  Xu, Li, and Zhao}]{zhang2026shortopd}
Zhang, Q.; Yuan, Q.; Lin, H.; Lu, Y.; Han, X.; Sun, L.; Li, X.; Xu, M.; Li, J.;
  and Zhao, X. 2026{\natexlab{b}}.
\newblock ShortOPD: Recovering Pruned LLMs with Short-to-Long On-Policy
  Distillation.
\newblock \emph{arXiv preprint arXiv:2607.13124}.

\bibitem[{Zhang et~al.(2026{\natexlab{c}})Zhang, Chai, Fu, Tu, Wang, Lin, Yin,
  Zhang, Zhu, and Zhao}]{zhang2026full}
Zhang, Y.; Chai, J.; Fu, Y.; Tu, S.; Wang, X.; Lin, W.; Yin, G.; Zhang, Q.;
  Zhu, Y.; and Zhao, D. 2026{\natexlab{c}}.
\newblock Are Full Rollouts Necessary for On-Policy Distillation?
\newblock \emph{arXiv preprint arXiv:2605.31490}.

\bibitem[{Zhao et~al.(2026)Zhao, Song, Tian, Shao, and Li}]{zhao2026prefix}
Zhao, Q.; Song, H.; Tian, S.; Shao, J.; and Li, X. 2026.
\newblock Prefix-Guided On-Policy Distillation: Mining Golden Trajectories from
  Rollouts.
\newblock \emph{arXiv preprint arXiv:2606.21994}.

\bibitem[{Ziheng et~al.(2026)Ziheng, Li, Tang, Wu, and
  Terzopoulos}]{ziheng2026less}
Ziheng, Z.; Li, J.; Tang, H.; Wu, Y.~N.; and Terzopoulos, D. 2026.
\newblock Less is more: Early stopping rollout for on-policy distillation.
\newblock \emph{arXiv preprint arXiv:2605.27028}.

\end{thebibliography}

\end{document}